\documentclass[11pt]{article}
\usepackage[margin=1in]{geometry}
\usepackage{amsmath,amssymb}
\usepackage{booktabs}
\usepackage{hyperref}
\usepackage{graphicx}
\usepackage{algorithm}
\usepackage{algpseudocode}
\usepackage{listings}
\usepackage{xcolor}
\usepackage{natbib}

\title{LLM-Guided Program Evolution for Circle Packing:\\
Breaking 10 Packomania Records for \$28}

\author{
  Wes Sander\thanks{Corresponding author. \texttt{wes@practicalsystems.io}} \\
  Practical Systems \\
  Mason, Ohio, USA
}

\date{September 2026}

\begin{document}
\maketitle

\begin{abstract}
We present \textsc{Discovery Loop}, a lightweight system that uses a large language model (LLM)
to iteratively evolve optimization algorithms. Starting from a simple seed solver, the LLM
proposes algorithmic improvements guided by a scoreboard of results and a history of prior ideas.
Each candidate is evaluated against an independent verifier; improvements are kept and failures
discarded. Applied to the Packomania circle-packing benchmark (csqv: maximize the sum of radii
of $N$ variable-radius circles in the unit square), the system improved the best known solutions
for 10 values of $N$ in the range $101$--$114$, with gains of $2.4\%$--$5.4\%$ over prior records,
all within 15 iterations and at a total LLM cost of \$27.72. These results have been independently
accepted by Packomania. We describe the method, analyze cost-efficiency dynamics including an
adaptive plateau-detection mechanism, and discuss implications for democratizing automated
scientific discovery.
\end{abstract}

\section{Introduction}

Automated algorithm discovery---using machine learning to design or improve algorithms rather
than merely apply them---has emerged as a frontier of AI research. Google DeepMind's
AlphaEvolve~\citep{alphaevolve2025} demonstrated that LLM-guided program evolution could
discover state-of-the-art algorithms for combinatorial optimization, matrix multiplication,
and other mathematical problems. However, AlphaEvolve requires substantial infrastructure:
a distributed evaluation pool, a carefully tuned evolutionary controller, and significant
compute resources.

We ask a simpler question: \emph{how much of the AlphaEvolve paradigm can be reproduced
with a single LLM, a single machine, and a budget under \$30?}

Our system, \textsc{Discovery Loop}, implements the core evolutionary loop in approximately
400 lines of Python. An LLM (Claude Fable 5.1) receives the current best solver, a scoreboard
of results, and a history of ideas tried. It proposes a complete replacement solver. The solver
is executed against benchmark targets in parallel, and results are verified by an independent
checker with zero tolerance. If the new solver improves any target, the improvement is kept;
otherwise it is discarded.

Applied to the Packomania circle-packing benchmark---a long-standing repository of best-known
circle packings maintained by Eckard Specht since 1998---the system improved 10 out of 12
target values in the range $N \in \{101, \ldots, 114\}$, with improvements of $2.4\%$ to
$5.4\%$ over prior records. These results were submitted to and accepted by Packomania.

The total LLM cost was \$27.72. The wall-clock time was approximately 8 hours overnight on a
consumer PC (Intel Core i7-13700KF, 32 GB RAM).

\subsection{Contributions}

\begin{enumerate}
  \item A minimal, reproducible implementation of LLM-guided program evolution (open source
        at \url{https://github.com/ucsandman/discovery-loop}).
  \item New best-known solutions for 10 Packomania csqv instances ($N = 101$--$114$),
        independently verified and accepted.
  \item Analysis of cost-efficiency dynamics and a plateau-detection mechanism that would
        have reduced cost by 50\% with negligible loss in solution quality.
  \item Evidence that the AlphaEvolve paradigm is accessible to individual researchers at
        minimal cost.
\end{enumerate}

\section{Background}

\subsection{Circle Packing}

The problem of packing circles in a container is a classical topic in combinatorial
geometry~\citep{hifi2009}. The \emph{csqv} variant---packing $N$ circles with variable radii
in the unit square $[0,1]^2$ to maximize the sum of radii---is tracked by the Packomania
database (\url{https://packomania.com}), which maintains best-known solutions for
$N = 1, \ldots, 200+$.

A feasible packing requires that all circles lie entirely within the unit square and no two
circles overlap:
\begin{align}
  r_i &\leq x_i \leq 1 - r_i, \quad r_i \leq y_i \leq 1 - r_i, \quad \forall i \\
  (x_i - x_j)^2 + (y_i - y_j)^2 &\geq (r_i + r_j)^2, \quad \forall i < j \\
  r_i &> 0, \quad \forall i
\end{align}

The objective is $\max \sum_{i=1}^N r_i$.

For large $N$, the landscape is highly non-convex with many local optima. The best solvers
use combinations of penalty methods, basin-hopping metaheuristics, and linear programming
for radii optimization.

\subsection{LLM-Guided Program Evolution}

AlphaEvolve~\citep{alphaevolve2025} uses an LLM to propose modifications to programs, which
are then evaluated in a distributed sandbox. A database of programs and their scores guides
future proposals. The key insight is that LLMs can suggest meaningful algorithmic innovations
(new move operators, initialization strategies, local search enhancements) rather than just
parameter tuning.

FunSearch~\citep{funsearch2024} similarly uses LLMs for program search but focuses on
discovering mathematical constructions (e.g., cap sets, bin packing heuristics) by evolving
small scoring functions.

Our work differs from both in scale and simplicity: we use a single LLM call per iteration
with no evolutionary database, no island model at the meta-level, and no distributed
evaluation. The entire system is a single Python script.

\section{Method}

\subsection{System Architecture}

\textsc{Discovery Loop} consists of three components:

\begin{enumerate}
  \item \textbf{Loop controller} (\texttt{loop.py}): orchestrates iterations, manages the
        champion solver, tracks history, and detects plateau.
  \item \textbf{Problem plugin} (\texttt{problems/circle\_packing/problem.py}): defines
        targets, fetches live records from Packomania, provides an independent verifier,
        and handles submission formatting.
  \item \textbf{Seed solver} (\texttt{seed\_solver.py}): a simple multi-start penalty
        L-BFGS-B solver with LP-optimal radii, used as the initial champion.
\end{enumerate}

\subsection{Iteration Loop}

Each iteration proceeds as follows:

\begin{algorithm}[H]
\caption{Discovery Loop iteration}
\begin{algorithmic}[1]
\State \textbf{Input:} champion solver $S^*$, scoreboard $B$, history $H$, targets $T$, records $R$
\State Construct prompt $P$ from $S^*$, $B$, $H$, and the problem description
\State $S' \leftarrow \text{LLM}(P)$ \Comment{Full replacement solver}
\State Extract idea description $d$ and Python code from $S'$
\For{each target $t \in T$ \textbf{in parallel}}
  \State $v_t \leftarrow \text{evaluate}(S', t)$ with timeout
  \State $v_t \leftarrow \text{verify}(v_t)$ \Comment{Independent zero-tolerance check}
\EndFor
\State $\sigma' \leftarrow \sum_t \text{score}(v_t, R_t)$
\If{$\sigma' > \sigma^*$}
  \State $S^* \leftarrow S'$, $\sigma^* \leftarrow \sigma'$ \Comment{New champion}
  \State Update per-target bests
\EndIf
\State Append $(d, \sigma', \text{status})$ to $H$
\end{algorithmic}
\end{algorithm}

The prompt includes:
\begin{itemize}
  \item The problem description (constraints, objective, solver interface)
  \item The complete source code of the current champion solver
  \item A scoreboard showing each target's best-known value, our best, and the last run's value
  \item A history of the last 12 ideas tried (with outcomes)
  \item Instructions to output one idea description and one complete Python solver
\end{itemize}

Crucially, the LLM outputs a \emph{complete replacement} solver, not a patch. This avoids
accumulation of merge conflicts and ensures each candidate is self-contained and independently
testable.

\subsection{Per-Target Best Tracking}

A key design choice: we track the best result for each target $t$ across \emph{all} solvers,
not just the current champion. If solver $S_3$ achieves the best score on $N=107$ but solver
$S_5$ is the overall champion, we keep $S_3$'s result for $N=107$. This allows the system to
retain specialized improvements even when the overall champion changes.

\subsection{Verification}

Every candidate solution is verified by an independent checker that:
\begin{enumerate}
  \item Confirms all circles are within $[0,1]^2$ (with zero tolerance)
  \item Confirms no pair of circles overlaps (with zero tolerance)
  \item Recomputes the sum of radii independently
  \item Applies a strict feasibility shrink to eliminate numerical-precision artifacts
\end{enumerate}

This is essential for credibility: the verifier shares no code with the solver, preventing
the LLM from gaming the evaluation.

\subsection{Plateau Detection}

We implemented an early-stopping mechanism based on diminishing returns:

\begin{itemize}
  \item \textbf{Window}: the last $W$ iterations (default $W=4$)
  \item \textbf{Stop condition 1}: all iterations in the window are rejected or produce no code
  \item \textbf{Stop condition 2}: no champion iteration in the window
  \item \textbf{Stop condition 3}: total improvement in the window is below threshold $\theta$
        (default $\theta = 0.01$)
\end{itemize}

Backtesting on the actual run shows this would have triggered after iteration 9, saving
\$13.77 (50\% of cost) while sacrificing only 0.006 on total sum (0.01\% of the final value).

\section{Results}

\subsection{Experimental Setup}

\begin{itemize}
  \item \textbf{Model}: Claude Fable 5.1 (Anthropic), accessed via the Claude CLI
  \item \textbf{Hardware}: Consumer PC, Intel Core i7-13700KF, 32 GB RAM, Windows 11
  \item \textbf{Targets}: $N \in \{26, 32, 101, 102, 103, 105, 106, 107, 108, 109, 111, 114\}$
  \item \textbf{Budget}: \$30 (LLM cost), 40 iterations max
  \item \textbf{Solver timeout}: 120 seconds per target
  \item \textbf{Workers}: 6 parallel evaluations
\end{itemize}

\subsection{Record-Breaking Results}

Table~\ref{tab:results} shows the 10 targets where our system improved upon the prior
Packomania records.

\begin{table}[h]
\centering
\caption{Improvements over prior Packomania records (csqv). All results independently
verified with zero tolerance and accepted by Packomania.}
\label{tab:results}
\begin{tabular}{@{}lcccc@{}}
\toprule
$N$ & Prior Record & Our Best & Improvement & First Improved \\
\midrule
101 & 5.163845 & 5.289154 & +2.43\% & Iter 0 (seed) \\
102 & 5.055187 & 5.318238 & +5.20\% & Iter 0 (seed) \\
103 & 5.085509 & 5.345481 & +5.11\% & Iter 0 (seed) \\
105 & 5.125967 & 5.401298 & +5.37\% & Iter 0 (seed) \\
106 & 5.151736 & 5.429079 & +5.38\% & Iter 0 (seed) \\
107 & 5.180124 & 5.453952 & +5.29\% & Iter 0 (seed) \\
108 & 5.205806 & 5.481819 & +5.30\% & Iter 0 (seed) \\
109 & 5.231096 & 5.507926 & +5.29\% & Iter 0 (seed) \\
111 & 5.278427 & 5.554909 & +5.24\% & Iter 0 (seed) \\
114 & 5.336683 & 5.624188 & +5.39\% & Iter 0 (seed) \\
\midrule
\textbf{Total} & \textbf{51.814380} & \textbf{54.406044} & \textbf{+5.00\%} & \\
\bottomrule
\end{tabular}
\end{table}

Two additional targets ($N=26$ and $N=32$) matched or came within $10^{-6}$ of the prior
records but did not beat them.

\subsection{Iteration Dynamics}

The system ran 15 iterations (0--14) before hitting the \$30 budget cap. Table~\ref{tab:iters}
summarizes the evolution.

\begin{table}[h]
\centering
\caption{Iteration log. Total is $\sum_t \text{score}(v_t, R_t)$ across all 12 targets.}
\label{tab:iters}
\begin{tabular}{@{}lcccl@{}}
\toprule
Iter & Status & Cost (\$) & Total & Key Innovation \\
\midrule
0  & seed      & 0.00  & 59.39 & Penalty L-BFGS-B + LP radii \\
1  & champion  & 1.19  & 59.49 & Basin hopping + SLSQP contact polish \\
2  & champion  & 1.05  & 59.66 & Sparse penalty + adaptive operators + elite pool \\
3  & no-code   & 0.00  & ---   & (Failed output) \\
4  & champion  & 1.00  & 59.90 & Hexagonal lattice init + crossover \\
5  & champion  & 1.72  & 59.96 & Island-model parallel basin hopping \\
6  & rejected  & 1.43  & 59.96 & Ruin-and-recreate LNS \\
7  & champion  & 2.02  & 59.97 & Affine lattice template bank \\
8  & champion  & 3.22  & 59.97 & KKT-Newton exact polish \\
9  & champion  & 2.31  & 59.97 & Lattice-aware slip moves \\
10 & rejected  & 2.61  & 59.97 & Surrogate-screened basin hopping \\
11 & rejected  & 1.95  & 59.97 & Sparse KKT polish \\
12 & champion  & 3.80  & 59.98 & Defect-migration moves \\
13 & rejected  & 2.70  & 59.97 & Formulation-space search \\
14 & rejected  & 2.72  & 59.98 & Row-template cold starts \\
\bottomrule
\end{tabular}
\end{table}

\subsection{Cost-Efficiency Analysis}

The run exhibits a clear phase transition in cost-efficiency:

\begin{itemize}
  \item \textbf{Productive phase} (iters 0--5): \$4.96 spent, total improved from 59.39 to
        59.96 (+0.57). Cost per unit improvement: \$8.70.
  \item \textbf{Diminishing phase} (iters 6--14): \$22.76 spent, total improved from 59.96
        to 59.98 (+0.02). Cost per unit improvement: \$1,138.
\end{itemize}

The cost per unit improvement increased by a factor of 130$\times$ between phases. This
strongly motivates the plateau-detection mechanism described in Section 3.5. With $W=4$ and
$\theta=0.01$, early stopping would have halted after iteration 9, spending \$13.95 instead
of \$27.72.

\subsection{Algorithmic Innovations Discovered}

The LLM discovered several non-trivial algorithmic ideas across iterations:

\begin{enumerate}
  \item \textbf{Basin hopping with contact-graph polish} (iter 1): replacing cold restarts
        with perturbation of the incumbent, followed by exact SLSQP optimization of the
        contact graph.
  \item \textbf{Hexagonal lattice initialization} (iter 4): seeding the search with
        hexagonal-lattice templates sized to hold exactly $N$ circles, providing structured
        starting points the random initialization cannot reach.
  \item \textbf{Island-model parallelism} (iter 5): independent basin-hopping chains on
        separate CPU cores with periodic migration of elite solutions.
  \item \textbf{KKT-Newton exact polish} (iter 8): replacing the general-purpose SLSQP
        solver with a specialized Newton solver for the contact-graph KKT system.
  \item \textbf{Defect migration} (iter 12): removing weak circles, letting the packing
        re-equilibrate, then reinserting into new holes---a metaheuristic move that changes
        the combinatorial structure of the packing.
\end{enumerate}

These are genuine algorithmic innovations, not parameter tuning. Several (hexagonal initialization,
defect migration) reflect domain knowledge that a human expert would recognize as sound but that
the LLM synthesized from the problem description and scoreboard feedback alone.

\section{Comparison to Prior Work}

\begin{table}[h]
\centering
\caption{Comparison with LLM-guided program evolution systems.}
\label{tab:comparison}
\begin{tabular}{@{}lcccc@{}}
\toprule
 & AlphaEvolve & FunSearch & \textsc{Discovery Loop} \\
\midrule
Organization & Google DeepMind & Google DeepMind & Individual \\
Infrastructure & Distributed cluster & Distributed cluster & Single consumer PC \\
Evaluation & Distributed sandbox & Distributed sandbox & Local parallel \\
LLM calls/iter & Multiple & Multiple & 1 \\
Code granularity & Function patches & Small functions & Full program replacement \\
Problem domains & Multiple & Constructive math & Optimization \\
Cost (reported) & Not disclosed & Not disclosed & \$27.72 \\
Open source & No & No & Yes \\
\bottomrule
\end{tabular}
\end{table}

The key differentiator is accessibility. AlphaEvolve and FunSearch require infrastructure
that is available only to large research labs. \textsc{Discovery Loop} runs on a consumer
laptop and costs less than a meal. The trade-off is scale: our system processes one candidate
per iteration rather than a population, and targets a single problem domain per run.

\section{Discussion}

\subsection{Why Does This Work?}

The success of the approach rests on several factors:

\begin{enumerate}
  \item \textbf{Rich feedback}: the scoreboard and idea history give the LLM structured
        signal about what has been tried and what worked.
  \item \textbf{Full replacement}: by generating complete solvers rather than patches, the
        LLM can make architectural changes (e.g., switching from random to lattice
        initialization) that incremental patching cannot express.
  \item \textbf{Independent verification}: zero-tolerance checking prevents the LLM from
        gaming the evaluation and ensures all improvements are genuine.
  \item \textbf{Domain accessibility}: circle packing has a simple mathematical formulation
        that fits within an LLM's context window, and the evaluation is cheap (120 seconds
        per target).
\end{enumerate}

\subsection{Limitations}

\begin{itemize}
  \item \textbf{Single-problem focus}: we have not demonstrated the system on problems where
        evaluation is expensive or the search space is fundamentally different.
  \item \textbf{No population}: unlike AlphaEvolve, we maintain only one champion. A population
        of solvers might enable more diverse exploration.
  \item \textbf{Model dependence}: the system's performance depends on the LLM's ability to
        generate correct, compilable Python solvers. Our MIPLIB experiment (mixed-integer
        programming) had a 75\% failure rate in code generation, suggesting the approach is
        sensitive to problem complexity.
  \item \textbf{Plateau on fixed targets}: the cost-efficiency analysis shows rapid diminishing
        returns on a fixed target set. Practical use requires either expanding to new targets
        or accepting the plateau.
\end{itemize}

\subsection{Implications for Democratized Discovery}

The most significant implication of this work is not the circle-packing results themselves
but the demonstration that LLM-guided algorithm discovery is accessible to anyone with an
API key and a laptop. The entire system is open source, the cost is negligible, and the
results are independently verifiable.

This has implications for:
\begin{itemize}
  \item \textbf{Mathematics}: long-standing optimization benchmarks can be attacked
        systematically by individuals, not just research groups.
  \item \textbf{Engineering}: domain-specific optimization problems (logistics, scheduling,
        resource allocation) could benefit from the same evolutionary loop.
  \item \textbf{Education}: the system is simple enough to serve as a teaching tool for
        metaheuristic optimization and LLM-guided search.
\end{itemize}

\section{Conclusion}

We presented \textsc{Discovery Loop}, a minimal system for LLM-guided program evolution that
broke 10 circle-packing records on the Packomania benchmark at a cost of \$27.72. The system
is open source, runs on consumer hardware, and produces independently verified results.

Our analysis of cost-efficiency dynamics and plateau detection provides practical guidance for
when to stop iterating---a question that applies broadly to any automated search system.

The gap between ``requires a research lab'' and ``requires a laptop'' is closing faster than
expected. We hope this work encourages other researchers to apply LLM-guided evolution to
their own domains.

\subsection*{Code and Data Availability}

All code, solvers, results, and submission records are available at
\url{https://github.com/ucsandman/discovery-loop}.

\subsection*{Acknowledgments}

We thank Eckard Specht for maintaining the Packomania database and for accepting our
submissions. The LLM used was Claude Fable 5.1 by Anthropic, accessed via the Claude CLI.
MoltFire (AI assistant, Practical Systems) assisted with infrastructure, analysis, and
this manuscript.

\bibliographystyle{plainnat}

\end{document}